\documentclass[runningheads]{llncs}
\usepackage[T1]{fontenc}
\usepackage{graphicx}
\usepackage{booktabs}
\usepackage{listings}%
\usepackage{multirow}
\usepackage{color}
\usepackage{colortbl}
\usepackage{paralist}
\usepackage{adjustbox}
\usepackage{multirow}
\usepackage{makecell}
\usepackage{amssymb}  
\usepackage{pifont}   
\usepackage{amsmath}
\definecolor{Di}{rgb}{0,0,0}

\newcommand{\xmark}{\ding{55}}  
\begin{document}
\title{From General Actions to Domain-Specific Monitoring: Prior-Adaptive Transfer for Skeleton-Based Action Recognition}

\titlerunning{Prior-Adaptive Transfer for Skeleton-Based Action Recognition}
%
\author{Hao Wang, Di Yang, Jiangtao Wang}
\authorrunning{}
%
\institute{University of Science and Technology of China}
\maketitle              
\begin{abstract}
Skeleton-based action recognition models have recently shown strong performance on large-scale benchmarks with general actions. However, directly transferring them to domain-specific tasks e.g., healthcare monitoring, is often suboptimal, as such tasks are narrow in scope and may be relevant to only a subset of general motion priors. 
Moreover, not all pretrained motion patterns are equally useful for a specific task, and retaining less relevant components may hinder adaptation and increase computational cost.
To address these challenges, we propose Prior-Adaptive Transfer of Skeletons (PATS), a framework that adapts general skeleton-based models by selectively retaining task-relevant motion priors while filtering redundant ones during transfer. PATS follows a standard pipeline that extracts skeleton signals from videos and employs a spatio-temporal backbone pre-trained on general actions. The key contribution lies in a novel Adaptive Prior Transfer module, which performs model compression as a prior selection mechanism through iterative pruning and refinement. Experiments on two specific action recognition tasks, Alzheimer’s detection and fall detection, show consistent improvements in both performance and efficiency over competitive baselines. The code will be released upon acceptance.

\keywords{Skeleton-based action recognition  \and Transfer learning \and Behavior analysis \and Pruning \and Prior knowledge.}
\end{abstract}
\section{Introduction}
\label{sec:intro}

\begin{figure*}[t]
  \centering
  \includegraphics[width=0.92\linewidth]{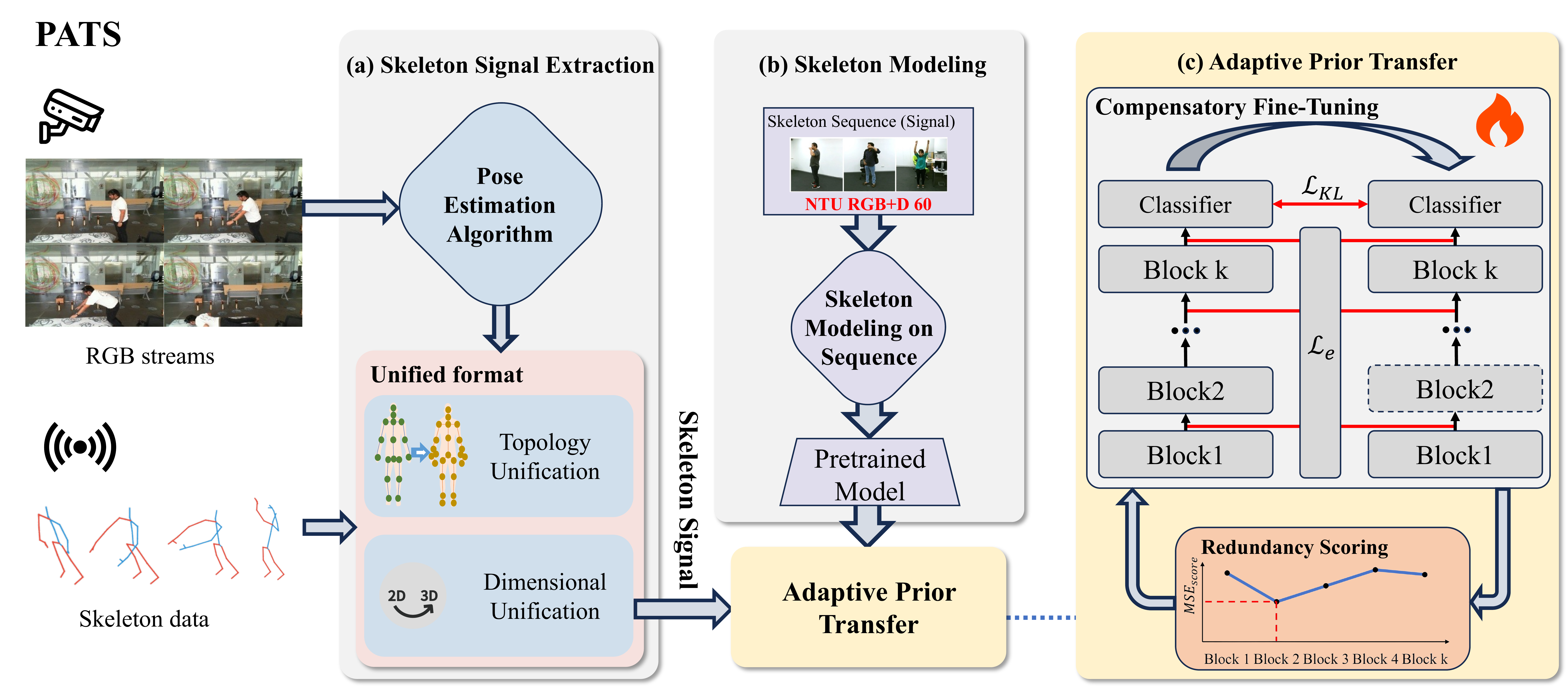}
  \vspace{-0.2cm}
  \caption{\textbf{Overview of the PATS framework.} 
  (a) Skeleton Signal Extraction: obtain standardized skeletons from raw data; 
  (b) Skeleton Action Modeling: encode spatio-temporal dynamics with a graph-based backbone; 
  (c) Adaptive Prior Transfer: adapt general action priors via pruning and fine-tuning for domain-specific tasks.}
  \label{fig:pipeline}
  \vspace{-0.2cm}
\end{figure*}

Human action recognition with skeleton sequences has achieved rapid progress, driven by large-scale datasets and increasingly sophisticated architectures ~\cite{arnab2021vivit,li2022uniformerv2,Yan2018SpatialTG,unik,topology_2021_ICCV,duan2021revisiting,do2025skateformer,yang2022via}. Owing to their robustness to background clutter and appearance variations, as well as their inherent advantages in privacy-preserving applications, skeleton sequences have shown great potential for a wide range of real-world applications. 
Despite these advances, most efforts emphasize architectural improvements and leaderboard performance, while the ability to generalize to low-data and domain-specific scenarios, e.g., healthcare monitoring~\cite{bruce2021skeleton,martinez2019up}, remains limited due to domain shifts and insufficient supervision.

Specifically, healthcare monitoring tasks often operate in small data regimes, where training task-specific models from scratch is infeasible due to the limited  data. Fortunately, pretrained skeleton action models can provide useful motion priors learned from general action datasets to support data-scarce downstream tasks. However, a domain gap often remains between the general action priors learned from large-scale skeleton action datasets and healthcare specific motion patterns, since healthcare scenarios typically involve subtle, clinically relevant movement deviations that are underrepresented in general action datasets.

Addressing this gap is critical for extending the impact of action recognition beyond controlled benchmarks. For example, in healthcare applications such as Alzheimer’s disease monitoring~\cite{bruce2021skeleton} or fall detection~\cite{martinez2019up}, systems must operate reliably and efficiently despite scarce annotated data and limited computational resources. 
This raises a key challenge for real world deployment: how can we effectively leverage the general action priors learned by large-scale pretrained models while improving algorithmic efficiency to satisfy the requirements of resource-constrained healthcare monitoring tasks?
This raises a key challenge for real-world deployment: how can we selectively leverage useful general action priors from large-scale pretrained models while removing less relevant components to improve efficiency for resource-constrained healthcare monitoring?

In this work, we propose \textit{\textbf{P}rior-\textbf{A}daptive \textbf{T}ransfer of \textbf{S}keletons (PATS)}, a unified framework that selectively retains target domain relevant motion priors from pretrained skeleton based recognition models and compresses the models for efficient adaptation to domain specific applications. PATS is composed of three modules: (a) a skeleton signal extraction module that produces standardized pose sequences from raw data, (b) a graph-based backbone that models spatio-temporal motion dynamics, and (c) an Adaptive Prior Transfer module that performs iterative pruning to retain blocks that are more useful for the target healthcare task while removing less relevant transformations, progressively pruning task irrelevant components and refining the remaining parameters. This design enables general action recognition models to generalize effectively in low-data domains while improving model inference efficiency. 
To the best of our knowledge, few prior studies have explored pruning in healthcare monitoring as a means of selecting target domain relevant priors from pretrained skeleton action models.

Our contributions can be summarized as follows:
\begin{itemize}
    \item[$\bullet$] We present PATS, a novel framework for adapting skeleton action models to domain-specific and low-data scenarios.
    
    \item[$\bullet$] We introduce an Adaptive Prior Transfer strategy that selectively transfers task-relevant motion priors from general models via redundancy-aware block pruning and compensatory adaptation.
    
    \item[$\bullet$] Our experimental analysis on Alzheimer's disease monitoring and fall detection using the EHE~\cite{bruce2021skeleton} and UP-Fall~\cite{martinez2019up} datasets demonstrates that PATS consistently improves over state-of-the-art methods.
\end{itemize}

\section{Related Work}
Graph-based networks such as ST-GCN~\cite{Yan2018SpatialTG} and InfoGCN~\cite{chi2022infogcn} explicitly encode human body topology and temporal dependencies, achieving state-of-the-art performance on large-scale benchmarks such as NTU RGB+D~\cite{Shahroudy2016NTURA,NTU-120}, which mainly contain general human actions. These pretrained representations can therefore be regarded as general motion priors learned from large-scale human motion data. In this work, we adopt InfoGCN as the backbone and investigate how to selectively transfer pretrained motion priors according to the requirements of healthcare-specific downstream tasks.

Transfer learning has been widely used to adapt pretrained skeleton action models to downstream tasks~\cite{alzubaidi2021novel,zhuang2020comprehensive}. Existing studies mainly rely on full parameter fine-tuning or parameter efficient adaptation strategies. However, healthcare monitoring tasks often involve domain-specific motion patterns that differ from general daily actions. Direct fine-tuning does not explicitly assess whether each pretrained motion prior is relevant to the target task. As a result, it may retain redundant or less relevant priors, which can limit adaptation effectiveness.

Model compression techniques, such as knowledge distillation~\cite{gou2021knowledge} and pruning~\cite{alzubaidi2021novel}, are important for resource constrained healthcare applications. Existing pruning methods mainly focus on generic model compression. Reconstruction-based pruning methods, such as ThiNet~\cite{luo2017thinet}, use feature reconstruction error as an effective criterion to identify redundant structures and guide filter pruning. In addition, BlockDrop~\cite{wu2018blockdrop} shows that residual blocks can be selectively skipped during inference to improve computational efficiency. However, less attention has been paid to using pruning as a transfer mechanism to remove task irrelevant or potentially interfering pretrained priors in healthcare monitoring tasks. This motivates our Adaptive Prior Transfer strategy, which selectively preserves useful motion transformations and removes redundant blocks for healthcare specific skeleton action recognition.
\vspace{-0.2cm}
\section{Method}

In this section, we present the proposed prior-adaptive transfer learning framework for skeletons (PATS). 
Section 3.1 presents an overview of the proposed framework, which is composed of three modules. Section 3.2 then describes in detail how iterative pruning is employed to selectively preserve target domain relevant motion priors.

\vspace{-0.1cm}
\subsection{Framework Overview}
As shown in Fig.~\ref{fig:pipeline}, PATS consists of three modules: (a) Skeleton Signal Extraction, (b) Skeleton Action Modeling, and (c) Adaptive Prior Transfer.

\begin{figure}[t]
  \centering
  \includegraphics[width=.7\linewidth]{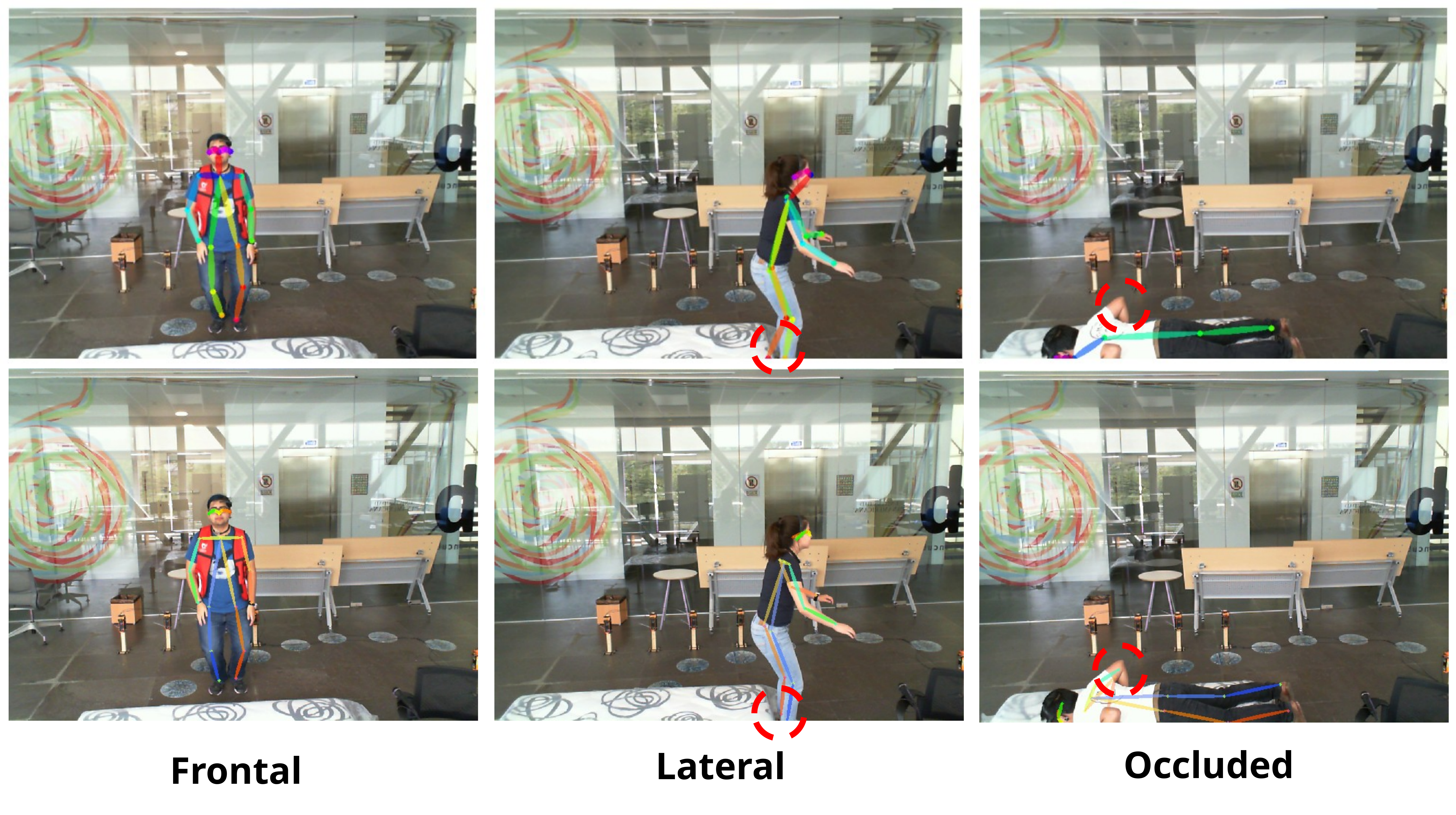}
  
\vspace{-0.3cm}

\caption{Qualitative comparison of OpenPose and AlphaPose on the UP-Fall dataset under \textbf{frontal}, \textbf{lateral}, and \textbf{occluded} views. AlphaPose shows more robust performance, particularly in challenging cases, such as those highlighted in the \textcolor{red}{\textbf{red circles}}.}
\vspace{-0.2cm}

\label{fig:skeleton_extraction}
\end{figure}

As shown in Fig.~\ref{fig:pipeline}(a), PATS first obtains skeleton data directly from sensors or extracts skeletons from RGB streams, providing compact and privacy-preserving motion representations. 
In PATS, AlphaPose is adopted as the default pose estimator for skeleton sequence extraction, owing to its robustness under lateral-view and occlusion scenarios, as illustrated in Fig.~\ref{fig:skeleton_extraction}. We further perform topological and dimensional unification to ensure consistency across datasets. Specifically, the extracted skeletons are mapped to the same topology as the NTU RGB+D dataset and standardized into a 25-joint representation.

As shown in Fig.~\ref{fig:pipeline}(b), the skeleton modeling module encodes the extracted skeleton sequences using InfoGCN~\cite{chi2022infogcn}, a graph-based backbone designed to capture both body topology and temporal dynamics. 
Although effective on benchmark datasets, direct transfer may be limited by distribution shifts. In PATS, the backbone thus acts as both a spatio-temporal encoder and a bridge, to expose general motion priors for adaptation by the Adaptive Prior Transfer module.

General action recognition datasets~\cite{Shahroudy2016NTURA,NTU-120} provide rich motion priors but differ significantly from task-oriented actions. 
To bridge this gap while improving efficiency, we introduce an Adaptive Prior Transfer module, as shown in Fig.~\ref{fig:pipeline}(c), which adapts pre-trained representations to health monitoring tasks and achieves model compression from the perspective of network depth. 
Before pruning, we first adapt the source-pretrained model to the target task, obtaining \(M_0\). This step projects general action priors into the target-domain distribution, ensuring that the subsequent redundancy estimation is guided by task characteristics.

\vspace{-0.1cm}
\subsection{Adaptive Prior Transfer}
\label{sec:adaptive}
In this section, we detail the underlying mechanism of Adaptive Prior Transfer. Healthcare monitoring tasks require models to achieve high recognition accuracy while maintaining computational efficiency~\cite{martinez2019up}.
Some research has demonstrated that Progressive Block Drop ~\cite{chen2025temporal} can effectively compress large-scale models, such as VideoMAE-S and VideoMAE-L~\cite{tong2022videomae}. Inspired by this, we explore its applicability to graph-based action recognition models.
Specifically, the Adaptive Prior Transfer module adopts an iterative pruning strategy to selectively retain useful general action priors from the pretrained model. This process consists of two key steps: \textit{Redundancy Scoring}, which estimates block-level redundancy using the training set and removes the most redundant block, and \textit{Compensatory Fine-Tuning}, which adapts the pruned model to compensate for the performance degradation caused by pruning.

\vspace{-0.2cm}

\subsubsection{Redundancy Scoring}
As for a block, we treat it as a fundamental computational unit within the network, comprising a sequence of operations responsible for transforming feature representations, such as a Transformer block in Transformer-based models.
By using blocks as the basic units for pruning, we can systematically evaluate and remove components that contribute least to target-task feature transformation, thereby selecting transferable general action priors while compressing the model for improved efficiency.

In this part, our objective is to identify the most redundant block in the current model, namely the block whose removal results in negligible or only marginal degradation in model performance. Specifically, at the \(n\)-th iteration, let \(M_{n-1}\) denote the model to be pruned, and let \(B_{n-1} = \{b_1, b_2, \ldots, b_K\}\) represent its \(K\) constituent blocks. Then, we perform a forward inference on the training data and compute the Mean Squared Error (MSE) between the block’s input and output feature representations. 
We use the input-output feature discrepancy as a proxy for estimating the transformation contribution of each block. 
Intuitively, if a block produces an output that is very close to its input on target-domain samples, the block performs only a weak transformation for the target task and can be viewed as less necessary after adaptation.
A smaller discrepancy suggests that the block behaves closer to an identity mapping under the target-domain distribution, and is therefore considered a candidate for pruning
~\cite{luo2017thinet,wu2018blockdrop,chen2026simdiff}. Formally, the evaluation criterion is defined as:

{\small
\begin{equation}
R_k =
\frac{1}{|\mathcal{D}_t|}
\sum_{x \in \mathcal{D}_t}
\left\|
\phi^{out}_{k}(x) - \phi^{in}_{k}(x)
\right\|_2^2,
\qquad
k^{*} = \arg\min_{k \in \{1,\ldots,K\}} R_k .
\end{equation}
}


where $D_t$ denotes the target domain training set, $\phi^{\mathrm{in}}_{k}(x)$ and $\phi^{\mathrm{out}}_{k}(x)$ represent the input and output features of the $k$-th block. The index $k^{*}$ corresponds to the block with the lowest MSE, which is identified as redundant and pruned.

\vspace{-0.1cm}

\subsubsection{Compensatory Fine-Tuning}
Although pruning removes redundant transformations, it may still perturb the learned feature distribution. As a result, pruning without adaptive fine-tuning typically degrades model performance~\cite{yeom2021pruning,le2021network}, making post-pruning optimization essential.
In our experiments, we adopt two strategies for Compensatory Fine-Tuning: full-parameter fine-tuning (FFT) and parameter efficient fine-tuning (PEFT) based on LoRA.
Experiments show that FFT achieves the highest accuracy, while LoRA provides efficiency benefits.
Therefore, LoRA is adopted to address scenarios with stricter efficiency requirements, such as edge computing deployments in community settings, while FFT is more suitable for accuracy oriented scenarios. 

Formally, \(M_n(\theta_n)\) denotes the model obtained after the \(n\)-th pruning iteration, where \(\theta_n\) represents the model parameters. Full-parameter fine-tuning is formulated as:
{\small
\begin{equation}
M^{F}_{n} = FT(M_n(\theta_n), D_t).
\end{equation}
}

For LoRA, low-rank matrices $\theta_{\mathrm{LoRA}}$ are inserted into the query and key projection layers of each block in InfoGCN, enabling task-specific adaptation with minimal parameter overhead:
{\small
\begin{equation}
M^{L}_{n} = FT(M_n(\theta_n+\theta_{LoRA}), D_t), \quad \theta_n \leftarrow \theta_n + \theta_{LoRA}.
\end{equation}
}

\paragraph{Inter-block Model Alignment:}
For action recognition, feature-level supervision is applied alongside the original training loss to recover performance of the pruned model~\cite{chen2025temporal}. For three consecutive blocks \(\{b_{i-1},b_i,b_{i+1}\}\), when the redundant block \(b_i\) is removed, the input to \(b^p_{i+1}\) in the pruned model is replaced by the output of \(b^p_{i-1}\), which is aligned with the corresponding output \(b^u_{i-1}\) from the unpruned model. Since \(b_i\) is the most redundant block, its output in the original model is nearly identical to that of \(b_{i-1}\). Block-level alignment is applied to all blocks against the unpruned model \(M_0\), yielding the inter-block alignment loss:

{\small
\begin{equation}
\mathcal{L}_{e} = \frac{1}{I-1} 
\sum_{m=1, m \neq i}^{I} 
\left\| e_{b_{m}}^{M_{0}} - e_{b_{m}}^{M_{n}} \right\|_{2}^{2}.
\end{equation}
}

where $e_{b_m}^{M_0}$ represents the output features of the $m$-th block in the unpruned model $M_0$, and $I$ denotes the total number of blocks in $M_0$.

\paragraph{Action-level Alignment:}
Besides the standard training loss, a Kullback–Leibler (KL) divergence loss \cite{chen2025temporal} aligns the class predictions:
{\small
\begin{equation}
\mathcal{L}_{KL} = D_{KL}(p(z_{M_0}) || p(z_{M_n})),
\end{equation}
}
where, \(z_{M_0}\) and \(z_{M_n}\) denote the logits of the unpruned and pruned models, and \(p(\cdot)\) denotes the softmax scores.

Accordingly, the total loss, incorporating the action recognition loss \(\mathcal{L}_{AR}\), can be expressed as:
{\small
\begin{equation}
\mathcal{L} = \mathcal{L}_{AR} + \mathcal{L}_{e} + \mathcal{L}_{KL}.
\end{equation}
}

\vspace{-0.2cm}
\section{Experiments and Analysis}

We evaluate PATS on two domain-specific low-data tasks: Alzheimer’s detection using the \textbf{EHE dataset}~\cite{bruce2021skeleton}, and fall detection using the \textbf{UP-Fall dataset}~\cite{martinez2019up}. As these tasks require distinct disease-related knowledge, e.g., Alzheimer’s manifests through subtle gait and motor irregularities, while fall detection involves sudden, high-risk motion patterns, these domains can provide a comprehensive evaluation of PATS under heterogeneous and challenging conditions. 

\noindent\textbf{Experimental Setting:}
To learn general action priors, we pretrain the backbone on the \textbf{NTU-RGB+D 60 dataset}~\cite{Shahroudy2016NTURA} following the cross-subject protocol, and then adapt the model to target tasks. The EHE dataset contains six categories of real morning exercise actions performed by individuals with Alzheimer's disease and healthy controls. Unlike general action recognition tasks that focus on action classification, the objective of this dataset is to determine whether a subject has Alzheimer's disease based on action performance. Therefore, the key challenge lies in identifying disease-related abnormalities from subtle intra class motion differences within the same type of action. In the experiments, we adopt a 5-fold cross-validation protocol, where each fold includes samples from different action categories. As for the UP-Fall dataset, it is designed for recognizing fall-related activities and daily actions, comprising five fall actions and six activities of daily living, for a total of eleven classes. Based on these benchmarks, the following subsections present our experimental results and analyses.

\subsection{Effect of Prior Knowledge}
In this section, we assess the impact of prior knowledge. Experimental results on the EHE and UP-Fall datasets show that models trained from scratch, without prior action knowledge, yield inferior performance. By contrast, NTU RGB+D 60 pretrained models with full parameter fine-tuning(FFT) consistently achieve superior results, as shown in Table~\ref{tab:the_impact_of_prior}, demonstrating the critical role of action prior knowledge in healthcare monitoring tasks.

\begin{table}[t]
\centering
\caption{Effect of Prior Knowledge on Downstream Tasks}
\label{tab:the_impact_of_prior}
\vspace{-0.1cm}

\resizebox{0.68\columnwidth}{!}{%
\begin{tabular}{l c ccccc c}
\toprule
\multirow{2}{*}{\textbf{Method}} &
\multicolumn{5}{c}{\textbf{EHE}} &
\multirow{2}{*}{\textbf{UP-Fall}} \\
\cmidrule(lr){2-6}
 & Fold1 & Fold2 & Fold3 & Fold4 & Fold5 &  \\
\midrule
Scratch 
& 72.41 & 69.86 & 78.89 & 64.76 & 68.09 & 90.00 \\

NTU60-Pretrained/FFT 
& \textbf{84.14} & \textbf{75.34} & \textbf{82.22} & \textbf{86.67} & \textbf{92.55} & \textbf{97.69} \\
\bottomrule
\end{tabular}
}
\vspace{-0.1cm}
\end{table}

\subsection{Adaptive Prior Knowledge Transfer}
\label{sec:compression}

Most existing studies adapt pretrained skeleton action models to downstream tasks by simply fine-tuning all pretrained components on the target datasets,  as analyzed in Section~4.1. However, healthcare monitoring tasks often involve domain-specific motion patterns such as subtle movement deviations, abnormal postural dynamics and clinically relevant action variations. These patterns may only require a limited subset of the general action priors learned from large-scale pretraining. Therefore, large-scale pretrained skeleton action models may contain redundant priors for healthcare-specific tasks, which can hinder effective adaptation and introduce unnecessary computational cost.

Our experiments provide empirical evidence for this redundancy. Simply retaining all pretrained components may preserve priors that are less relevant or even distracting for the target healthcare domain. As a result, it can lead to suboptimal recognition accuracy and increased computational overhead. By contrast, the proposed Adaptive Prior Transfer strategy selectively preserves task-relevant priors and removes redundant components. The resulting compressed model achieves performance comparable to or sometimes better than the uncompressed model while offering improved efficiency.

First, we compare the uncompressed model with compressed models obtained using different compression methods for removing redundant priors on the EHE dataset. To ensure a fair comparison, all models are initialized from the target adapted pretrained model $M_0$, which serves as the uncompressed baseline.
Table \ref{tab:compression_method} compares our adaptive prior transfer strategy with two common compression baselines: model distillation and weight pruning. Both baselines degrade performance relative to the uncompressed model. For distillation, the student model removes the 3rd, 6th, and 9th blocks while preserving the remaining structure. A logits-based distillation loss is introduced to align the student and teacher output distributions via KL divergence with a temperature of 2. For weight pruning, we prune 30\% of the model parameters based on the L1-norm magnitude, resulting in approximately 1.079M parameters after pruning.

\begin{table}[t] 
\centering 
\caption{Comparison of Different Model Compression Strategies for Adaptive Transfer on the EHE and UP-Fall.
PATS (LoRA) denotes the configuration in which LoRA is used for compensatory fine-tuning.
Prune\# denotes the number of pruned blocks.} 
\label{tab:compression_method} 
\vspace{-0.1cm} 
\resizebox{0.8\columnwidth}{!}{%
\begin{tabular}{l c ccccc c} 
\toprule 
\multirow{2}{*}{\textbf{Method}} & 
\multirow{2}{*}{\textbf{Prune \#}} & 
\multicolumn{5}{c}{\textbf{EHE}} & 
\multirow{2}{*}{\textbf{UP-Fall}} \\ 
\cmidrule(lr){3-7} & & Fold1 & Fold2 & Fold3 & Fold4 & Fold5 & \\ 
\midrule 

NTU60-Pretrained/FFT & -- & 84.14 & 75.34 & 82.22 & 86.67 & 92.55 & 97.69 \\ 

Distillation & -- & 68.28 & 80.14 & 82.22 & 84.29 & 92.55 & 91.54 \\ 

Weight-Pruning & -- & 80.56 & 66.44 & 80.56 & 85.24 & 91.49 & 97.69 \\ 

\midrule 
\multirow{3}{*}{PATS (LoRA)} 
& 1 & 84.83 & 78.08 & 80.00 & 84.76 & 91.49 & 93.08 \\ 
& 2 & 82.07 & \textbf{80.14} & 79.44 & 84.29 & 90.43 & 93.85 \\ 
& 3 & 79.31 & 77.40 & 77.78 & 84.76 & 86.17 & 90.00 \\ 

\midrule 
\multirow{3}{*}{PATS (FFT)} 
& 1 & 85.52 & 78.08 & \textbf{85.56} & \textbf{87.62} & \textbf{93.62} & \textbf{98.46} \\ 
& 2 & \textbf{86.90} & 78.77 & 85.00 & 86.67 & 93.09 & \textbf{98.46} \\ 
& 3 & 83.45 & 77.40 & 85.00 & 84.29 & 93.09 & 97.69 \\ 
\bottomrule 
\end{tabular} } 
\end{table}

\begin{table*}[ht]
\centering
\caption{Comparison of Model Efficiency on the EHE and UP-Fall Datasets in Terms of Latency, GFLOPs, and Parameters}
\label{tab:pruned_latency_ehe_upfall}

\resizebox{0.88\textwidth}{!}{%
\begin{tabular}{l c c cccccc}
\toprule

\multirow{2}{*}{\textbf{Method}} & 
\multirow{2}{*}{\textbf{Prune \#}} & 
\multirow{2}{*}{\textbf{Metric}} & 
\multicolumn{5}{c}{\textbf{EHE}} & 
\multirow{2}{*}{\textbf{UP-Fall}} \\

\cmidrule(lr){4-8}
& & & 
\textbf{F1} & \textbf{F2} & \textbf{F3} & \textbf{F4} & \textbf{F5} & \\

\midrule

\multirow{3}{*}{NTU60-Pretrained/FFT}
& \multirow{3}{*}{--}
& Lat.(ms)
& 56.76 & 54.84 & 55.60 & 55.19 & 56.24 & 57.77 \\
& & GFLOPs
& 1.66 & 1.66 & 1.66 & 1.66 & 1.66 & 1.66 \\
& & Params(M)
& 1.542 & 1.542 & 1.542 & 1.542 & 1.542 & 1.544 \\

\midrule
\multirow{9}{*}{PATS (LoRA)}
& \multirow{3}{*}{1}
& Lat.(ms)
& 55.62 & 53.24 & 53.32 & 52.66 & 52.17 & 56.95 \\
& & GFLOPs
& 1.38 & 1.59 & 1.38 & 1.59 & 1.38 & 1.38 \\
& & Params(M)
& 1.185 & 1.517 & 1.185 & 1.517 & 1.185 & 1.187 \\

& \multirow{3}{*}{2}
& Lat.(ms)
& 54.02 & 52.75 & 52.98 & 51.19 & 51.29 & 55.45 \\
& & GFLOPs
& 1.31 & 1.31 & 1.31 & 1.31 & 1.31 & 1.31 \\
& & Params(M)
& 1.160 & 1.160 & 1.160 & 1.160 & 1.160 & 1.162 \\

& \multirow{3}{*}{3}
& Lat.(ms)
& 52.43 & 52.12 & 51.95 & 50.47 & 50.68 & 53.42 \\
& & GFLOPs
& 1.16 & 1.23 & 1.23 & 1.16 & 1.16 & 1.16 \\
& & Params(M)
& 1.068 & 1.135 & 1.135 & 1.068 & 1.068 & 1.071 \\

\midrule
\multirow{9}{*}{PATS (FFT)}
& \multirow{3}{*}{1}
& Lat.(ms)
& 55.54 & 52.73 & 54.22 & 54.48 & 54.56 & 56.09 \\
& & GFLOPs
& 1.38 & 1.59 & 1.38 & 1.59 & 1.38 & 1.38 \\
& & Params(M)
& 1.185 & 1.517 & 1.185 & 1.517 & 1.185 & 1.187 \\

& \multirow{3}{*}{2}
& Lat.(ms)
& 53.33 & 52.32 & 53.87 & 53.43 & 52.20 & 55.51 \\
& & GFLOPs
& 1.31 & 1.52 & 1.31 & 1.52 & 1.31 & 1.31 \\
& & Params(M)
& 1.160 & 1.492 & 1.160 & 1.492 & 1.160 & 1.162 \\

& \multirow{3}{*}{3}
& Lat.(ms)
& 51.84 & 50.43 & 51.53 & 51.89 & 51.74 & 54.01 \\
& & GFLOPs
& 1.16 & 1.37 & 1.16 & 1.37 & 1.16 & 1.16 \\
& & Params(M)
& 1.068 & 1.401 & 1.068 & 1.401 & 1.068 & 1.071 \\

\bottomrule
\end{tabular}
}
\end{table*}

As for our pruning method, following the principle of the adaptive prior transfer strategy described in Section\ref{sec:adaptive}, we begin by transferring a general model pretrained on NTU-RGB+D 60 to the Alzheimer's detection task, where the uncompressed model is fine-tuned on each fold of the EHE dataset. Redundant blocks are then pruned, and compensation is performed through fine-tuning with either FFT or LoRA. The corresponding results are reported in Table \ref{tab:compression_method}. Our method achieves the best performance, demonstrating the effectiveness of the proposed approach. This suggests that selectively transferring general action priors while removing redundant components enables the model to outperform the uncompressed baseline, particularly when FFT is applied. 

Regarding efficiency, the efficiency improvements achieved by our method are reported in Table~\ref{tab:pruned_latency_ehe_upfall}. In the experiments, we compared the models from three aspects: latency, GFLOPs, and the number of model parameters. According to the results, our method provides approximately a 10\% improvement in latency, thereby enhancing the model’s efficiency in real-world scenarios. Additionally, since the number of model parameters is reduced by around 30\% in most cases, the method can overcome storage limitations and broaden the potential deployment scenarios in practical applications.

According to the experimental setting, for the Alzheimer’s detection task, which requires determining Alzheimer’s based on a person performing a specific action. Therefore, both the uncompressed and compressed models are further fine-tuned with full-parameter updates on each action category, allowing them to better capture the diagnostic mechanisms of Alzheimer’s disease and thereby improve model performance. As shown in Table \ref{tab:sota_ehe_upfall}, both the uncompressed and compressed models consistently outperform the SOTA across nearly all categories, highlighting the value of prior action knowledge and the importance of more fine-grained adjustments of the model for each action category. Moreover, when comparing the uncompressed and compressed models, the compressed models generally achieve superior performance, with the best results obtained under FFT. 
The results in Table \ref{tab:sota_ehe_upfall} represent the averages from 5-fold cross-validation across all action categories.

Furthermore, we conducted comparative experiments on the UP-Fall dataset, with the results presented in Table \ref{tab:compression_method} and Table \ref{tab:sota_ehe_upfall}. Consistent with our previous findings, the results suggest that the pretrained model contains components that are less necessary for the target task.
Pruning effectively removes such redundancy, reducing the model size while maintaining or even improving recognition performance. As a result, the pruned model achieves the best overall results.
In addition, as shown in Table~\ref{tab:pruned_latency_ehe_upfall}, the model also achieves improvements in both inference latency and parameter count. Therefore, in different task scenarios, our method consistently achieves strong performance, further demonstrating its effectiveness.

\begin{table}[t]
\centering
\caption{Comparison with previous methods on EHE and UP-Fall. The best results are indicated in bold, and the second-best results are underlined.}
\label{tab:sota_ehe_upfall}

\resizebox{0.85\columnwidth}{!}{%
\begin{tabular}{l c ccccccc c}
\toprule
\multirow{2}{*}{\textbf{Method}} & 
\multirow{2}{*}{\textbf{Prune \#}} & \multicolumn{6}{c}{\textbf{EHE}} & \multirow{2}{*}{\textbf{UP-Fall}} \\
\cmidrule(lr){3-8}
 & & E1 & E2 & E3 & E4 & E5 & E6 \\
\midrule
\multicolumn{1}{l}{Previous SOTA~\cite{bruce2021skeleton,martinez2019up}}       & -- & 95.37 & 82.59 & 79.72 & 90.72 & 66.43 & 69.21 & 95.10 \\
\multicolumn{1}{l}{Uncompressed} & -- & 96.99 & 89.56 & 95.12 & \underline{94.25} & 85.06 & 87.02 & \underline{97.69} \\
\midrule
\multirow{3}{*}{PATS (LoRA)} 
 & 1 & \underline{97.09} & 90.22 & 95.02 & 91.94 & 84.82 & 88.28 & 93.08 \\
 & 2 & 95.49 & 90.30 & 94.12 & 91.46 & 84.46 & 85.45 & 93.85 \\
 & 3 & 95.83 & 88.96 & 96.64 & 92.71 & 85.06 & 84.12 & 90.00 \\
\midrule
\multirow{3}{*}{PATS (FFT)} 
 & 1 & \textbf{97.15} & \underline{93.11} & \underline{97.71} & \textbf{94.72} & \textbf{87.32} & 86.94 & \textbf{98.46} \\
 & 2 & 94.87 & \textbf{93.26} & \textbf{97.88} & 93.00 & \underline{85.71} & \underline{89.80} & \textbf{98.46} \\
 & 3 & 94.18 & 88.22 & 96.64 & 94.15 & 84.64 & \textbf{90.21} & \underline{97.69} \\
\bottomrule
\end{tabular}
}
\end{table}

\begin{table}[!t]
\centering
\caption{Ablation study on different loss functions. The results correspond to models with a single block pruned, followed by full parameter fine-tuning.}
\label{tab:ablation}
\vspace{-0.1cm}

\resizebox{0.6\columnwidth}{!}{%
\begin{tabular}{cc*{5}{c}}
\toprule
\multirow{2}{*}{$\mathcal{L}_{KL}$} & \multirow{2}{*}{$\mathcal{L}_{e}$} & \multicolumn{5}{c}{Fold Results (\%)} \\
\cmidrule(lr){3-7}
 & & Fold 1 & Fold 2 & Fold 3 & Fold 4 & Fold 5 \\
\midrule
\checkmark & \checkmark & 85.52 & 78.08 & 85.56 & 87.62 & 93.62 \\
\xmark & \checkmark & 84.83\textcolor{red}{$_{\downarrow0.69}$} & 67.81\textcolor{red}{$_{\downarrow10.27}$} & 85.00\textcolor{red}{$_{\downarrow0.56}$} & 84.29\textcolor{red}{$_{\downarrow3.33}$} & 92.55\textcolor{red}{$_{\downarrow1.07}$} \\
\checkmark & \xmark & 74.48\textcolor{red}{$_{\downarrow11.04}$} & 64.38\textcolor{red}{$_{\downarrow13.70}$} & 81.11\textcolor{red}{$_{\downarrow4.45}$} & 80.48\textcolor{red}{$_{\downarrow7.14}$} & 89.36\textcolor{red}{$_{\downarrow4.26}$} \\
\xmark & \xmark & 81.38\textcolor{red}{$_{\downarrow4.14}$} & 64.38\textcolor{red}{$_{\downarrow13.70}$} & 81.11\textcolor{red}{$_{\downarrow4.45}$} & 83.81\textcolor{red}{$_{\downarrow3.81}$} & 88.83\textcolor{red}{$_{\downarrow4.79}$} \\
\bottomrule
\end{tabular}%

}
\vspace{-0.2cm}
\end{table}

\vspace{-0.2cm}

\subsection{Ablation Study}
\noindent\textbf{Effect of Loss Functions:}
To assess the effects of the additional inter-block alignment loss $\mathcal{L}_{e}$ and the action-level alignment loss $\mathcal{L}_{KL}$ incorporated into the Adaptive Prior Transfer strategy, we conduct experiments on the EHE dataset, with the results summarized in Table~\ref{tab:ablation}. Specifically, after pruning a single block, we apply full parameter fine-tuning for compensatory adaptation and evaluate the resulting model. As shown in Table~\ref{tab:ablation}, we examine the impact of removing $\mathcal{L}_{e}$, removing $\mathcal{L}_{KL}$, and removing both losses simultaneously. The results demonstrate that both the inter-block alignment loss and the action-level alignment loss are important for recovering the performance of the pruned model.

\begin{figure}[t]
  \centering
  \includegraphics[width=.8\linewidth]{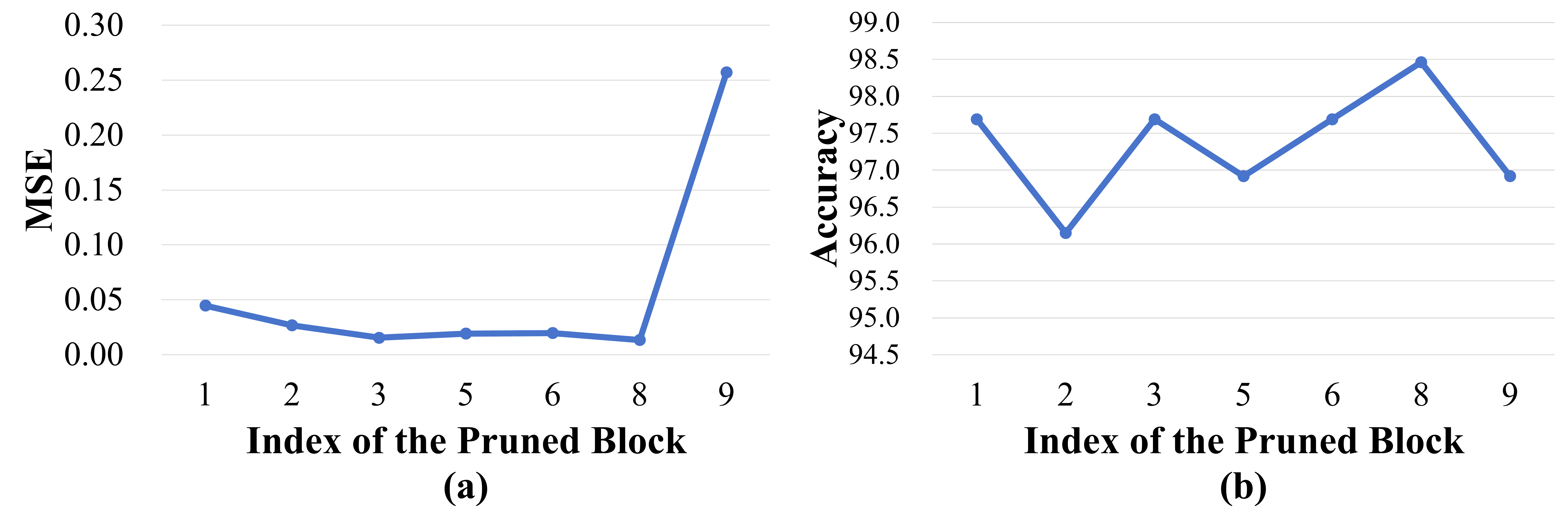}
  
\vspace{-0.3cm}

\caption{Ablation on Pruned Block Index. (a) MSE \textit{v.s.} different block;(b) Accuracy \textit{v.s.} different block}

\label{fig:mse_ablation}
\end{figure}

\noindent\textbf{Redundancy Evaluation Criteria MSE:}
As shown in Fig.\ref{fig:mse_ablation}(a), based on the UP-Fall dataset, we use MSE to measure the redundancy of each block, where a smaller MSE indicates a smaller impact of the block on feature transformation and thus higher redundancy. In the ablation study, we evaluate the performance of compressed models after pruning blocks with varying MSE values. As for Blocks 4 and 7, since they perform feature dimension transformations with different input and output dimensions, they are not considered for pruning during compression. The experimental results are presented in Fig.\ref{fig:mse_ablation}(b), showing that the compressed model achieves the best performance when Block 8 with the smallest MSE is removed. This verifies the effectiveness of MSE as a metric for measuring block redundancy.

\vspace{-0.2cm}

\section{Conclusion}
We presented PATS, a prior-adaptive transfer framework that enables skeleton-based action recognition models to generalize from large-scale benchmarks to low-data healthcare tasks. Experiments on low-data scenarios show that transferring broad action priors, combined with prior-guided pruning, yields both accuracy and efficiency. These findings highlight the potential of leveraging general motion knowledge for robust deployment in domain-specific applications.

\vspace{-0.1cm}

\bibliographystyle{splncs04}
\bibliography{mybibliography}

\end{document}